\def\year{2020}\relax
\documentclass[letterpaper]{article} 
\usepackage{aaai20}  
\usepackage{times}  
\usepackage{helvet} 
\usepackage{courier}  
\usepackage[hyphens]{url}  
\usepackage{graphicx} 
\usepackage{graphicx}  
\usepackage{subfigure}
\usepackage{algpseudocode}
\usepackage{algorithm}
\usepackage{mathtools}
\algnewcommand\algorithmicforeach{\textbf{for each}}
\algdef{S}[FOR]{ForEach}[1]{\algorithmicforeach\ #1\ \algorithmicdo}
\usepackage{amsmath}
\usepackage{amssymb}
\usepackage{comment}
\title{Improving the Evaluation of Generative Models with Fuzzy Logic}
\author{Julian Niedermeier\thanks{Equal contribution.}, Gon\c{c}alo Mordido\footnotemark[1], Christoph Meinel\\
Hasso Plattner Institute\\ 
Prof.-Dr.-Helmert-Straße 2-3\\
14482 Potsdam, Germany\\
julian.niedermeier@student.hpi.uni-potsdam.de, goncalo.mordido@hpi.de, christoph.meinel@hpi.de 
}
 
\begin{document}

\maketitle

\begin{abstract}
Objective and interpretable metrics to evaluate current artificial intelligent systems are of great importance, not only to analyze the current state of such systems but also to objectively measure progress in the future. In this work, we focus on the evaluation of image generation tasks. We propose a novel approach,  called Fuzzy Topology Impact (FTI), that determines both the quality and diversity of an image set using topology representations combined with fuzzy logic. When compared to current evaluation methods, FTI shows better and more stable performance on multiple experiments evaluating the sensitivity to noise, mode dropping and mode inventing.
\end{abstract}

\section{Introduction}

Accurate evaluation of a model's learning capabilities is of extreme importance to identify possible shortcomings in the model's behavior. 
When learning a discriminative, supervised task, this evaluation is often straightforward by comparing the model's predictions against ground-truth labels. For example, in an image classification task with labeled data, one can evaluate the model's label prediction of an image on the test set to its real label.

However, in a generative, unsupervised task, the assessment of a model's capabilities is far more challenging. As an example, considering image generation with unlabeled data using generative adversarial networks~\cite{gans}, a model would generate an image from random noise. How can one evaluate the quality of such an image? Moreover, how can one evaluate the diversity of the entirety of the generated set? Answering these questions is the focus of this work.

Our method builds on top of the topological representations created by UMAP's algorithm~\cite{umap}. These topological features can be represented by a directed, weighted graph which first uses the k-nearest neighbors (KNN) algorithm to establish the connections between nodes. Then, such connections are weighted using principles of Riemannian geometry and fuzzy logic, representing the probability of the existence of each directed edge in the resulting graph. 

We call our evaluation method Fuzzy Topology Impact (FTI), that has as basis the construction of two of the aforementioned graphs, one for the real and one for the fake data. Then, we analyze the impact that each sample of a given set has on the other set's graph to separately determine the quality and diversity of the fake data set. More precisely, quality is measured by the impact, on average, that a fake sample has on the real data graph, and diversity is measured inversely, by measuring the impact each real sample has on the fake data graph.

In the end, our method can be interpreted as the drop in the average probability of the existence of a connection in the real graph and fake graph, representing the quality and diversity of the fake data. We present the following contributions:
\begin{enumerate}
    \item Retrieval of two interpretable metrics, which directly correlate to sample quality and diversity. 
    \item Contrarily to previous topology-based methods, our method can be seen as finer-grained approach due to the usage of fuzzy logic.
    \item Thorough experimental discussion of existing evaluation methods, \textit{i.e. } Inception Score~\cite{inception_score}, Fréchet Inception Distance~\cite{fid}, precision and recall assessment~\cite{sajjadi2018assessing}, and improved precision and recall~\cite{kynkaanniemi2019improved}, showing the superiority of our approach.
    \item Code for the reproducibility of the results is available at \url{https://github.com/sleighsoft/fti}.
\end{enumerate}

\section{Related Work}

This work primarily focuses on the evaluation of image generation models
targeting the evaluation of both image quality and diversity. In general, current approaches can be categorized into three different types: analysis of likelihoods~\cite{theis2015note} and probability distributions~\cite{fid,gretton2012kernel}, topological analysis of manifolds~\cite{sajjadi2018assessing,kynkaanniemi2019improved,khrulkov2018geometry}, and classifier-based methods~\cite{inception_score,gurumurthy2017deligan,shmelkov2018good}. This work falls within the topological analysis category, where we propose a novel approach that improves existing metrics by following a finer-grained methodology. A description of the methods compared throughout this paper follows.


Inception score or IS~\cite{inception_score} analyzes the output distribution of a pre-trained Inception-V3~\cite{szegedy2016rethinking} on ImageNet~\cite{imagenet} to measure both the quality and diversity of a fake image set. To this end, they use the Kullback-Leibler Divergence to compare the conditional probability distribution of a fake sample being classified as a given class as well as the marginal distribution of all samples across the existing classes. Higher IS should indicate that each fake sample is clearly classified as belonging to a single class and that all fake samples are uniformly distributed across all existing classes.

Fréchet Inception Distance or FID~\cite{fid} builds upon the idea of using the Inception-V3 network, but this time to simply obtain feature representations. FID, in contrast to IS, uses the real data distribution and retrieves a distance to the fake data distribution. Therefore, a lower FID is better since it indicates the fake distribution approximates the real one. Even though FID provides significant improvements over IS, like the detection of intra-class mode dropping where only identical images of each class are generated, it also retrieves a single-valued metric. Therefore, it does not give a direct insight regarding the quality and diversity of the generated set.

To fix this, \cite{sajjadi2018assessing} proposed to separate the evaluation into two distinct values, namely precision and recall, by using the relative probability densities of the real and fake distributions. For simplicity, we refer to this approach as Precision and Recall for Distributions (PRD). Thus, precision reflects the quality of generated images, whereas recall quantifies the diversity in the fake image set. Using Inception-V3's features, similarly to FID, for both real and fake samples, they use k-means clustering to group the totality of the samples and evaluate quality and diversity by analyzing the histograms of discrete distributions over the clusters' centers for the real and fake data. Precision and recall values are approximated by calculating a weighted F-Score with $\beta = 8$ and $\beta = \dfrac{1}{8}$, respectively.

Having concerns about how to appropriately choose $\beta$ and reliability against mode dropping or truncation, \cite{kynkaanniemi2019improved} proposed to use non-parametric representations of the manifolds of both real and fake data. We refer to this approach as IMproved Precision And Recall (IMPAR). Instead of using Inception-V3, IMPAR uses VGG-16~\cite{simonyan2014very}'s feature representations. Moreover, instead of determining a set of clusters in the data, as proposed by PRD, IMPAR uses KNN to approximate the topology of the underlying data manifold by forming a hypersphere to the third nearest neighbor of each data point. Precision is then the fraction of points in the fake image set that lie within the real data manifold, whereas recall is the fraction of points in the real image set that lie within the generated data manifold.

Since IMPAR uses a binary overlapping approach to compare the real and fake data manifolds, it lacks into taking into consideration sample density. For example, when dealing with highly sparse data, big regions of the data space may intersect - think of a binary overlapping version of Figure~\ref{fig:neighbors_3_uniform}. This may also be observed when using a high $K$.
In this work, we propose a finer-grained, mathematical sound KNN approach based on fuzzy logic that is sensitive to different overlapping regions depending on the overall sample density. 

\section{Fuzzy Topology Impact}

Following the method proposed by UMAP~\cite{umap}, we create a graph where each node represents the embeddings from a pre-trained model of each image. The resulting weighted, directed graph is designed to maintain the topological representations of the embeddings using Fuzzy logic, with each weight representing the \textit{probability of the existence} of a given edge. Then, we measure the drop in the average probability of existence that a new sample has in the original graph, which we call the Fuzzy Topology Impact (FTI). Following this principle, we separately analyze the quality, by calculating the impact that fake samples have in the real samples' graph, and diversity, by measuring the impact that real samples have in the fake samples' graph.

\subsection{Topological Representation}
\label{sec:topological_representation}

We will now dive into the underlying properties used by UMAP that enable the data manifold approximation with a fuzzy simplicial set representation in the form of a weighted graph. The geodesic distance from a given point to its neighbors can be normalized by the distance of the k-th neighbor (or by a scaling factor $\sigma$), creating a notion of local distance that is different for each point. This notion aligns with the assumption that the data is uniformly distributed on the manifold with regards to a Riemannian metric (see \cite{umap} for original lemmas and proofs), which is a requirement for the theoretical foundations from Laplacian eigenmaps~\cite{belkin2002laplacian,belkin2003laplacian} used to formally justify this manifold approximation.

When combining the aforementioned principles with Riemannian geometry, most concretely by connecting each data point using 1-dimensional simplices, we achieve a weighted, directed, k-neighbor graph that represents the approximated manifold. The weight values of the resulting graph are computed using fuzzy logic, which inherently describes the probability of the existence of each edge.

Given $N$ embeddings, $X = \{x_1, \ldots, x_N\}$, and the $k \in \mathbb{N}$ nearest neighbors under the euclidean distance $d \in \mathbb{R}^+$ of each $x_i \in X$, $\{x_{i_{1}}, \ldots, x_{i_{k}}\}$, we have the following graph $G$: $G = (V, E)$, where $V$ represents the embeddings $X$ and $E$ forms a set of directed edges, $E \subseteq \{(x_{i}, x_{i_{j}}) \mid j \in \mathbb{N}: j \in [1, k] \land i \in \mathbb{N}: i \in [1, N]\}$. Each directed edge $e_{x_i, x_{i_{j}}} \in E$, is associated with the following weight or probability of existence $p_{x_i, x_{i_{j}}} \in \mathbb{R}^+: p_{x_i, x_{i_{j}}} \in [0,1]$:

\begin{equation}
    p_{x_i, x_{i_{j}}} = \exp\bigg(\dfrac{-d(x_{i}, x_{i_{j}})}{\sigma_i}\bigg),
    \label{eq:probability_existence}
\end{equation}

where $\sigma_i \in \mathbb{R}^+_*$ represents the scaling factor associated with $x_i$ such that:

\begin{equation}
    \sum_{j=1}^{k} \exp\bigg(\dfrac{-d(x_{i}, x_{i_{j}})}{\sigma_i}\bigg) = \log_2(k).
    \label{eq:sigma}
\end{equation}

Thus, the existence probability associated with each embedding's connections are scaled such that the cardinality of the resulting fuzzy set is fixed: $\sum_{j=1}^{k} p_{x_i, x_{i_{j}}} = \log_2(k)$. Note that $\log_2(k)$ was chosen through an empirical search by the original UMAP implementation and we re-use this value. Such scaling standardizes the weights of the resulting graph while still maintaining the notion of local connectivity by the usage of individual scaling factors for each embedding.

\begin{figure}%
\centering
\subfigure[Original samples.]{%
\includegraphics[width=0.48\linewidth]{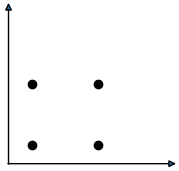}}%
\subfigure[Original weighted, directed graph.]{%
\includegraphics[width=0.48\linewidth]{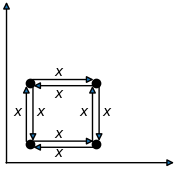}}%
\caption{Given a set of original samples represented as filled circles (a), we generate a weighted, directed graph using $k=2$ (b). Since all samples' closest neighbors are at the same distance, the same weight is shared among all edges.}
\label{fig:method}%
\end{figure}

The resulting graph is weighted and directed, with the corresponding weights representing the probability of existence of the directed connection between a point and respective neighbors. Figure~\ref{fig:method} provides a simple illustration of these principles.

Note that there are several differences between our final graph and UMAP's. While we use a directed graph, UMAP combines disagreeing weights to represent the probability of at least one of the edges existing to form an undirected graph. Contrarily to UMAP, we set the local connectivity to 0, meaning that the weight of each sample's closest neighbor is not set to $1.0$. This was done to mitigate the influence of outliers in the retrieved impact. Moreover, each node in the graph represents each sample's embeddings from a pre-trained model instead of the sample itself. We found using the embedding information to be more stable in our experiments. Finally,  instead of finding a low dimensional representation from the resulting graph, we use the inherent topological information to evaluate generative models, which is described next.

\subsection{Impact Evaluation}
\label{sec:evaluation_phase}

Considering the previously described graph $G$, we can calculate the average probability of existence of the directed edges by:

\begin{equation}
 \overline{P_G} =\dfrac{\sum_{i=1}^{N} \sum_{j=1}^{k} p_{x_i, x_{i_{j}}}}{N \times k}.
\end{equation}

The proposed evaluation metric is to simply retrieve the average drop of $\overline{P_G}$ when adding a new sample $x^\prime_i$ to the original graph. To achieve this, we modify each weight in the following way:

\begin{equation}
\begin{split}
p^{x^\prime_i}_{x_i, x_{i_{j}}} =
    \begin{dcases*}
              0 ,\quad \text{if }  j = k \wedge d(x_i, x_{i_{k}}) > d(x_i, x^\prime_i)\\
              e^{\dfrac{-d(x_{i}, x_{i_{j}})}{\sigma^\prime_i}} ,\text{if }  j \neq k \wedge d(x_i, x_{i_{k}}) > d(x_i, x^\prime_i)\\
              p_{x_i, x_{i_{j}}} ,\quad \text{otherwise.}\\
            \end{dcases*}
\end{split}
\label{eq:graph_new}
\end{equation}

\begin{figure}%
\centering
\subfigure[Effects of a new, realistic sample on the original graph.]{%
\label{fig:method_real}%
\includegraphics[width=0.48\linewidth]{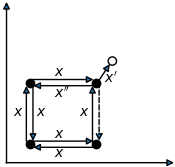}}%
\subfigure[Effects of a new, similar sample on the original graph.]{%
\label{fig:method_graph}%
\includegraphics[width=0.48\linewidth]{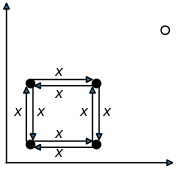}}%
\caption{Original samples are represented by filled circles whereas new samples are shown as empty circles. New samples that are the $k$ closest neighbor to a given original point will affect the weights of all directed edges from such point (a). Outlier samples, \textit{i.e.} new samples that are not a closest $k$ neighbor to any original point, cause no impact in the original graph (b).}
\label{fig:method_2}%
\end{figure}

Hence, if a new sample $x^\prime_i$ is part of the $k$ closest neighbors of an original sample $x_i$, we remove the connection to the original $k$'th furthest neighbor, \textit{i.e.} $p^{x^\prime_i}_{x_i, x_{i_{k}}} = 0$, and update the weight values of the original $k-1$ nearest neighbors according to Eq.~\ref{eq:probability_existence} and the new $\sigma^\prime_i$ satisfying Eq.~\ref{eq:sigma_new}. On the other hand, if $x^\prime_i$ is not a $k$ cloest neighbor to any original sample $x_i$, the original weight values remain unchanged. Figure~\ref{fig:method_2} illustrates these scenarios.

\begin{equation}
    \sum_{j=1}^{k-1} \Bigg(\exp\bigg(\dfrac{-d(x_{i}, x_{i_{j}})}{\sigma^\prime_i}\bigg)\Bigg) + \exp\bigg(\dfrac{-d(x_{i}, x^\prime_i)}{\sigma^\prime_i}\bigg) = \log_2(k).
    \label{eq:sigma_new}
\end{equation}

Thus, the drop of average probability of existence of the original connections by a new sample $x^\prime_i$ can be described as:

\begin{equation}
 \overline{P_{G,x^\prime_i}} =\dfrac{\sum_{i=1}^{N} \sum_{j=1}^{k} p^{x^\prime_i}_{x_i, x_{i_{j}}}}{N \times k}.
\end{equation}

Finally, having $X$ as the original set used to generate $G$ with $k$ nearest neighbors, and $N^\prime$ new samples $X^\prime = \{x_1^\prime, \ldots, x^\prime_{N^\prime}\}$, FTI can be defined as the average drop of probability of existence of the original connections:

\begin{equation}
 FTI(X, X^\prime, k) = \dfrac{\sum_{i=1}^{N^\prime} \overline{P_G} - \overline{P_{G, x^\prime_i}}}{N^\prime}.
\end{equation}

Algorithm~\ref{alg:method} provides a more practical view of the proposed method. Note that the presented pseudo-code is optimized for visualization, not performance.
The function $SmoothDistApprox$ executes a binary search that satisfies Equation~\ref{eq:sigma_new} for the distances passed as argument, similarly to UMAP.

\begin{algorithm} \caption{Fuzzy Toplogy Impact. $G$ represents the original graph and $dist$ a dictionary with the euclidean distances of each sample's nearest neighbors.}
\begin{algorithmic}[1]
\Require $X$, the original set of samples; $X^\prime$, the new set of samples; $k$, the number of neighbors
\State $impact \gets 0$
\ForEach {$x^\prime_i \in X^\prime $}
\State $p^X \gets 0$
\State $p^{X^\prime} \gets 0$
\State $count \gets 0$
\ForEach {$x_i \in X $}
\If{$d(x_i, x^\prime_i) < d(x_i, x_{i_{k}})$}
\State $count \gets count + 1$
\State $p^X \gets p^X + p_{x_i, x_{i_{k}}}$
\State \textbf{del} $dists[(x_i, x_{i_{k}})]$
\State $p^{x^\prime_i}_{x_i, x_{i_{k}}} \gets 0$
\State $dists[(x_i, x^\prime_i)] \gets d(x_{i}, x^\prime_i)$
\State $\sigma^\prime_i \gets SmoothDistApprox(dists, k)$
\For {$j=1, \ldots, k-1$}
\State $p^X \gets p^X + p_{x_i, x_{i_{j}}}$
\State $p^{x^\prime_i}_{x_i, x_{i_{j}}} \gets \exp\bigg(\dfrac{-d(x_{i}, x_{i_{j}})}{\sigma^\prime_i}\bigg)$
\State $p^{X^\prime} \gets p^{X^\prime} + p^{x^\prime_i}_{x_i, x_{i_{j}}}$
\EndFor
\EndIf
\EndFor
\State $impact \gets impact + p^X-p^{X^\prime}$
\EndFor
\State \textbf{return} $\dfrac{impact}{N^\prime}$
\end{algorithmic}
\label{alg:method}
\end{algorithm}



\subsubsection{Number of neighbors}

The open cover of the manifold is computed by finding the $k$-nearest neighbors of each original sample. Therefore, using smaller $k$ values promote a more detailed local structure, whereas larger $k$ values induce a larger, global structures. In another words, a higher number of neighbors leads to the resolution of which the
topology is approximated to become more diffused, spreading high impact over larger regions.

\begin{figure}%
\centering
\subfigure[$k=2.$]{%
\includegraphics[width=0.48\linewidth]{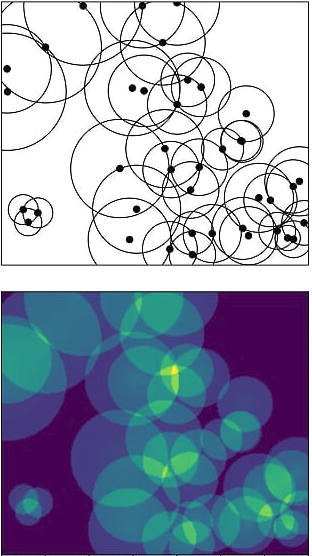}}%
\subfigure[$k=3$]{%
\includegraphics[width=0.48\linewidth]{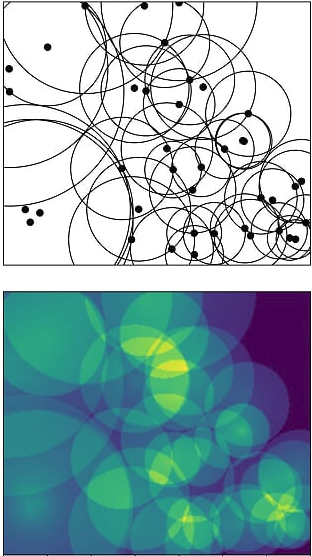}
\label{fig:neighbors_3_uniform}}%
\caption{Visualization of the impact of new points given randomly distributed original points using 2 (a) and 3 (b) neighbors. Warmer colors indicate higher impact than cooler colors, with the darkest color indicating no impact.}
\label{fig:neighbors_random}%
\end{figure}

To visualize such effect of using different number of neighbors in the overall impact, we analyze one toy example with 40 random original samples (Figure~\ref{fig:neighbors_random}). The top row shows the original samples in a 2-dimensional space with the radius to the k-th nearest neighbor, while the bottom row presents the impact a new sample would have at any given (x,y)-coordinate.


\subsubsection{Quality and Diversity}





We introduced FTI as the drop in the average probability of existence in the original graph. If we consider the real data as the original sample set $R$ and the generated data as the new sample set $G$, we can derive both the quality and diversity of the generated data by calculating the bi-directional impact between both sets. 

More specifically, quality can be defined as the impact that, on average, a fake sample has on the real data graph. In contrast, diversity is defined as the impact that, on average, a real sample has on the fake data graph. The two metrics are then defined as follows:

\begin{equation}
    quality = FTI(R, G, k)\qquad diversity = FTI(G, R, k)
\end{equation}

\section{Experimental Results}

We tested our approach alongside IS, FID, PRD, and IMPAR using three datasets: Fashion-MNIST~\cite{xiao2017fashion}, CIFAR-10 and CIFAR-100~\cite{krizhevsky2009learning}. The performed experiments evaluate the sensitivity to noise as well as the sensibility to mode dropping, mode addition and mode invention. 
Throughout our experimental setup, we used the training images and testing images of each dataset as real and generated samples, respectively. The embeddings used by our approach were calculated using Inception-V3 due to lower runtime than VGG-16. Since the different compared metrics have different ranges, we analyze the results using their respective ratios.

\subsection{Noise Sensitivity}

\begin{figure*}%
\centering
\subfigure{%
\includegraphics[width=0.33\linewidth]{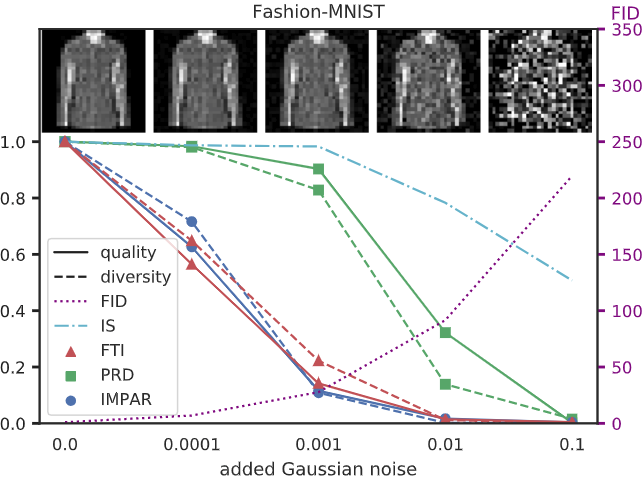}}%
\subfigure{%
\includegraphics[width=0.33\linewidth]{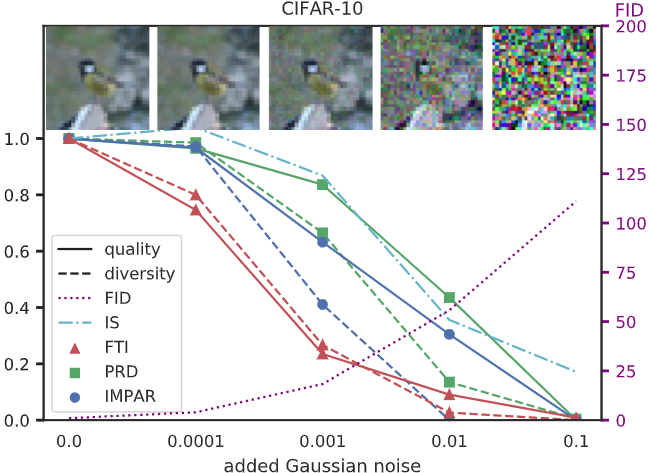}}%
\subfigure{%
\includegraphics[width=0.33\linewidth]{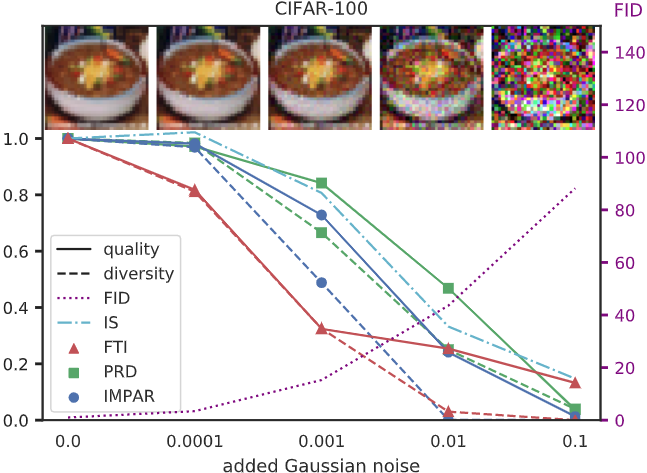}}%
\caption{Results for added Gaussian noise on Fashion-MNIST, CIFAR-10 and CIFAR-100. All metrics are normalized by their respective values obtained on unaltered test images, \textit{i.e. no added noise}.}
\label{fig:noise}%
\end{figure*}

To test the sensitivity of the different methods against different amounts of noise, we incrementally added Gaussian noise to the test images of each dataset. Ideally, all methods should show signs of deterioration and, while quality should decrease faster than diversity when little noise is added, both metrics should degrade. Figure~\ref{fig:noise} shows the comparison results.


We observe that FID is very sensitive to noise with distances growing by an order of magnitude even at almost imperceptible noise amounts. IS is barely perturbed by the noise on Fashion-MNIST and, unexpectedly, shows an increase on CIFAR-10 and CIFAR-100, as well as constant behavior at early noise stages on Fashion-MNIST. Similarly, PRD shows little sensitivity from low to mid noise amounts and then rapidly drops as noise increases. Even though IMPAR and FTI show similar performance, IMPAR shows a faster decrease in diversity over quality, which we argue is not ideal for this experiment. Finally, FTI shows the most levels of sensitivity which we directly link to the fine-grained property of our method.





\subsection{Mode Dropping}
\label{subsec:mode_dropping}

\begin{figure*}%
\centering
\subfigure{%
\includegraphics[width=0.33\linewidth]{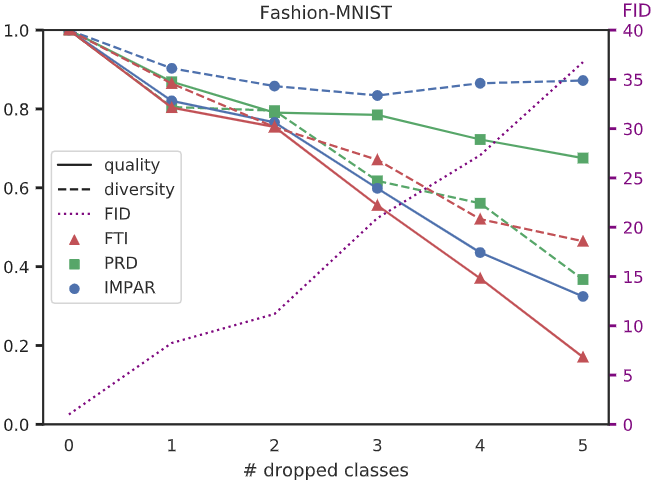}}%
\subfigure{%
\includegraphics[width=0.33\linewidth]{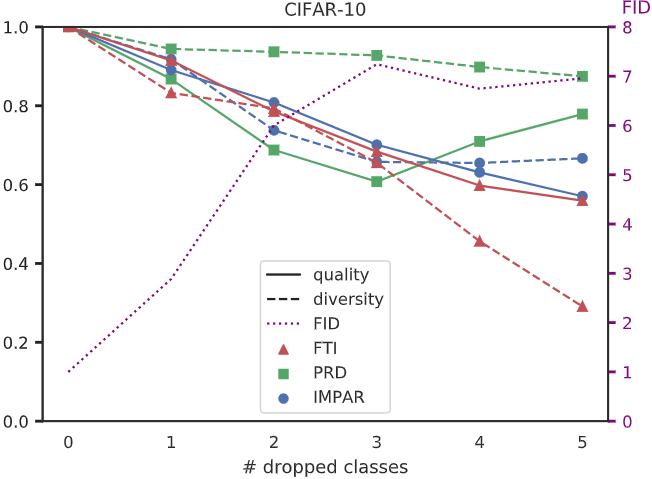}}%
\subfigure{%
\includegraphics[width=0.33\linewidth]{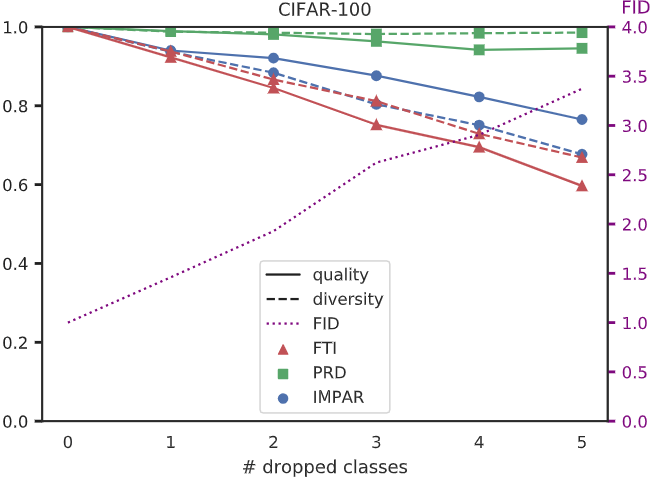}}%
\caption{Mode dropping results on Fashion-MNIST, CIFAR-10 and CIFAR-100. Metrics are normalized by their respective values on zero dropped classes.}
\label{fig:mode_dropping}%
\end{figure*}

We further simulated mode collapse by first defining a constant window that includes samples from only half of the classes of the different datasets as the real sample set. On the other hand, the test set window slides through the remaining classes, one class at a time, dropping samples from a class represented in the real sample set while adding samples from one unseen class. Ideally, all methods should show a proportional decrease with the number of real classes dropped. Moreover, quality is affected by adding samples from fake classes while diversity is also affected as real classes are removed from the test set. Figure~\ref{fig:mode_dropping} shows the comparison results. Note that IS is excluded from this experiment as it uses a pre-trained classifier on all classes.


We observe that FID almost linearly increases for Fashion-MNIST and CIFAR-100, but stagnates for CIFAR-10 at 3 dropped classes. PRD detects a change in the number of modes for Fashion-MNIST but does not capture mode dropping for CIFAR-10, as its quality first decreases and then increases unexpectedly, and CIFAR-100 where its decrease of both quality and diversity is negligible. IMPAR's diversity fails to detect a decrease in diversity on Fashion-MNIST, even showing an increase on CIFAR-10 when all classes are dropped. 
Overall, FTI is the most stable approach showing sensitivity to mode dropping across all datasets. 

\subsection{Mode Addition \& Invention}

\begin{figure*}%
\centering
\subfigure{%
\includegraphics[width=0.33\linewidth]{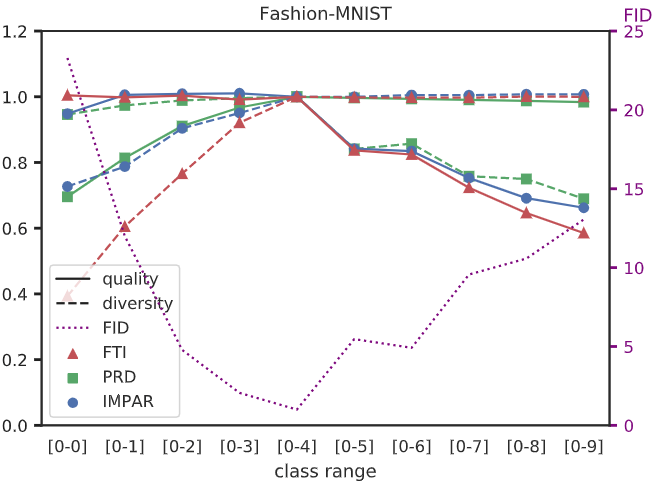}}%
\subfigure{%
\includegraphics[width=0.33\linewidth]{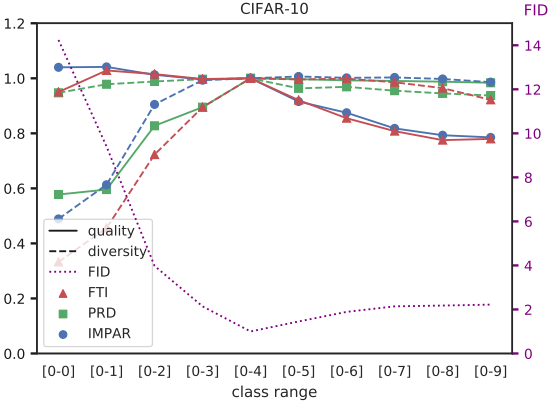}}%
\subfigure{%
\includegraphics[width=0.33\linewidth]{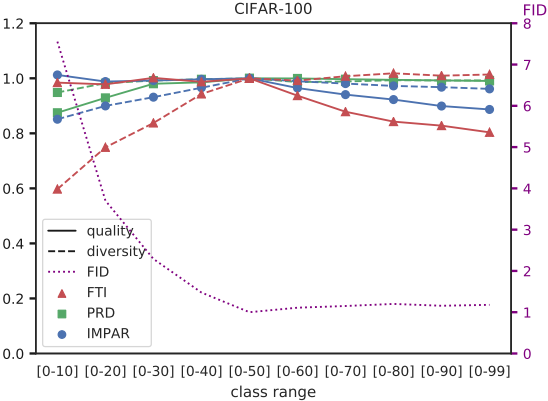}}%
\caption{Mode invention experiment on Fashion-MNIST, CIFAR-10, and CIFAR-100. Metrics are normalized by their respective values for [0-4], [0-4], and [0-50] class ranges, respectively.}
\label{fig:mode_inventing}%
\end{figure*}

We replicated \cite{sajjadi2018assessing}'s experimental setup to evaluate a different variant of mode collapse and inventing which sheds more light on the importance of using two separate metrics to measure quality and diversity independently. The window of the real set is identical to the last experiment, however, instead of a sliding window for the testing set, we simply add one class at a time, without dropping any class. 
Note that, since the cardinality of the test set changes, we do not normalize FTI by the number of original connections in this experiment.
Thus, this experiment measures mode addition until all real classes are present in the test set, and mode invention for additionally added classes. Ideally, the quality remains constant during the mode dropping phase, while diversity increases with each added class. In the mode invention phase, diversity should remain constant whereas quality should decrease as the added classes are not part of the real sample set. Figure~\ref{fig:mode_inventing} shows the comparison results.



On FID, we observe signs of sensitivity to mode collapse, as shown in the previous experiment, however, on CIFAR-10 and CIFAR-100, it fails to punish mode inventing with the overall distance remaining almost constant. Hence, we verify that FID's single-value is unclear with regards to image quality and diversity, as seen on Fashion-MNIST, reinforcing the importance of a separate analysis of quality and diversity. 
Nevertheless, PRD's quality and diversity behave contradictory to what is expected.
Moreover, on CIFAR-10, PRD's diversity stays constant which is also seen on CIFAR-100 for both quality and diversity. 
IMPAR assigns the same diversity to the class range [0-3] as it does to [0-4] for CIFAR-10 and it lacks to disentangle quality and diversity measures for CIFAR-100.
In conclusion, and once more, we see the expected behavior on FTI for this experiment, successfully detecting mode addition and mode invention across all data sets. 

\section{Conclusion and Future Work}

Accurately evaluating the performance of machine-generated content is of utmost importance. This work provides an in-depth look at four existing metrics on several experiments using three different datasets and multiple experiments concerning sensitivity to noise and detection of mode dropping and mode inventing. We propose a novel method that evaluates the quality and diversity of generated images using topological representations and fuzzy logic. The experimental results show the overall superiority of the proposed method as well as shortcomings of current approaches.

Since our method simply uses embedding information, it is not limited to image generation tasks solely. Thus, it would be interesting to test the effectiveness of our approach outside image generation, such as text generation tasks. In the future, we plan to extend our evaluation to real-world scenarios to solidify the proposed metrics.
    
\bibliography{biblio}
\bibliographystyle{aaai}

\end{document}